\documentclass[letterpaper, 10 pt, conference]{ieeeconf}  

\IEEEoverridecommandlockouts                             
\usepackage{graphics}
\usepackage{epsfig}
\usepackage{amsmath}
\usepackage{amssymb}
\usepackage{mathrsfs}
\usepackage{bm}
\usepackage{subfig}
\usepackage[noadjust]{cite}

\title{\LARGE \bf
Bounded Collision Force by the Sobolev Norm
}

\author{Kevin Haninger and Dragoljub Surdilovic
	\thanks{}
	\thanks{Authors affiliated with Fraunhofer Institut f{\"u}r Produktionsanlagen und Konstruktionstechnik (e-mail: \texttt{\{kevin.haninger, dragoljub.surdilovic\}@ipk.fraunhofer.de}). This project has received funding from the European Union’s Horizon 2020 programme under grant agreement No 820689}%
}

\begin{document}
	
\maketitle  

\begin{abstract}
	A robot making contact with an environment or human presents potential safety risks, including excessive collision force. While experiments on the effect of robot inertia, relative velocity, and interface stiffness on collision are in literature, analytical models for maximum collision force are limited to a simplified mass-spring robot model. This simplified model limits the analysis of control (force/torque, impedance, or admittance) or compliant robots (joint and end-effector compliance). Here, the Sobolev norm is adapted to be a system norm, giving rigorous bounds on the maximum force on a stiffness element in a general dynamic system, allowing the study of collision with more accurate models and feedback control. The Sobolev norm can be found through the $\mathcal{H}_2$ norm of a transformed system, allowing efficient computation, connection with existing control theory, and controller synthesis to minimize collision force. The Sobolev norm is validated, first experimentally with an admittance-controlled robot, then in simulation with a linear flexible-joint robot. It is then used to investigate the impact of  control, joint flexibility and end-effector compliance on collision, and a trade-off between collision performance and environmental estimation uncertainty is shown. 
\end{abstract}

\section{Introduction}
Mechatronic design for position control is largely standardized - every position-controlled robot has high drivetrain stiffness with a controller designed for high-bandwidth tracking performance and disturbance rejection. On the other hand, mechatronic design for physically interactive robots is more application-specific \cite{desantis2008, albu-schaffer2008,mosadeghzad2012}, where joint stiffness, end-effector compliance, and control architecture vary significantly. To design interactive robots, methods to quantify the impact of compliance and control on performance and safety are needed. Some trade-offs between performance and robustness have been shown \cite{valency2003}, and the role of joint compliance on relaxing coupled stability conditions investigated \cite{haninger2018a}, but unified design methods are not yet established.

\begin{table*}[t]
	\centering
	\renewcommand{\arraystretch}{1.2}
	\caption{Collision metrics in literature\label{collision_metrics}}
	\begin{tabular}{|r||l|l|l|}
		\hline 
		Quantity & Name & Domain & Notes, reference\tabularnewline
		\hline 
		\hline 
		$\max\left|f\left(t\right)\right|$ & Maximum Force{*} & Human, Environment & Most fundamental \cite{iso2016,vukobratovic2009,heinzmann2003,unfallversicherung2009}\tabularnewline
		\hline
		$f_{ss}=f(t)|_{t\rightarrow\infty}$ & Steady State Force & Clamped human, Stiffness environment & \cite{iso2016, surdilovic2007, huelke2012}\tabularnewline
		\hline
		$\max\left|f\left(t\right)v\left(t\right)\right|$ & Maximum Power & Human, Environment & Mentioned, but not investigated in \cite{iso2016}\tabularnewline
		\hline 
		$\int\left|f\left(t\right)v\left(t\right)\right|dt$ & Absolute Energy Transfer{*} & Human, Environment & Typ. taken over initial penetration \cite{vemula2018, povse2010, iso2016} \tabularnewline
		\hline 
		$\Delta_t^{-3/2}\left\Vert a\left(t\right)|\right\Vert _{2}^{5/2}$ & Head Injury Criterion & Human: Head, unconstrained &  2-norm only over $\Delta_t$, \cite{brands2002, haddadin2007} \tabularnewline
		\hline
	\end{tabular}
	\\ \vspace{3pt}
	$*$ Also extended to per-unit-area metrics (i.e. pressure, energy flux)
\end{table*}

One design requirement for interactive robots robot is establishing safe contact with a human or environment, in both collision (unintended) or contact transition (intended). To allow safe robots moving at reasonable speeds in semi-structured environments, a limit on contact force should be achieved over possible contacts. Design for contact is challenging, as contact with a stiff environment introduces broad frequency input to the robot, exciting high-frequency resonance modes which are often unmodelled. Additionally, the high-frequency dynamics of the robot can't be easily reshaped by control due to controller bandwidth limitations. This makes the intrinsic dynamics of a robot especially important in collision, motivating the need for physical compliance on otherwise high-impedance robots\cite{haddadin2007}.

\begin{figure}[t]
	\centering
	\includegraphics[width=\columnwidth]{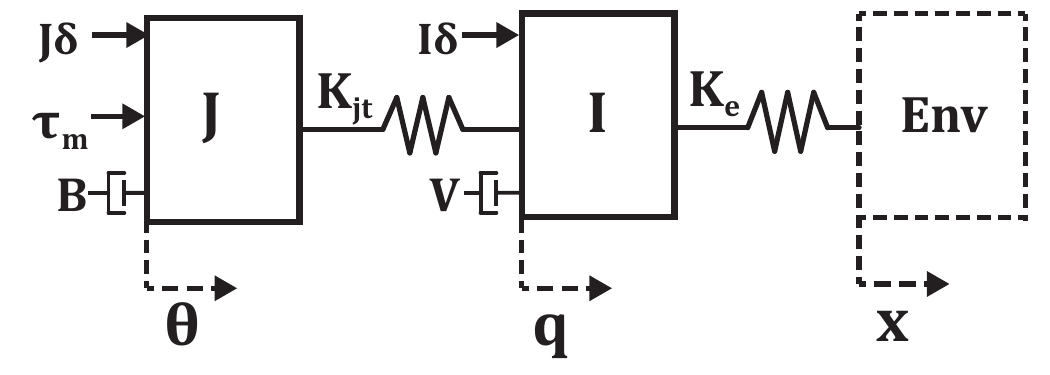} 
	\protect\caption{Two inertia robot model with motor inertia $J$, joint flexibility $K_{jt}$ (e.g. from torque sensor or harmonic gearbox), link inertia $I$, and compliance at the interface $K_e$ (e.g. soft skin, compliant end-effector), in contact with an environment. Scaled Dirac delta impulses $\delta$ induce the initial contact velocity.\label{physical_sys}}
\end{figure}

Compliance can be realized as a flexible joint between the motor and load \cite{albu-schaffer2007, pratt1995}, a compliant covering \cite{duchaine2009a}, or a soft end-effector. Choosing the stiffness of these compliant elements is as much an art as a science, relying on iterative testing, and realized values in commercial products are often kept as trade secrets.  Often, other important robot characteristics are fixed (e.g. motor inertia, link inertia), leaving the choice of stiffness and control architecture as important design choices.

Designing compliance and control to avoid user injury is important for human-robot applications. Characterizing human injury from robot collision is experimentally well-established, and several established metrics are shown in Table \ref{collision_metrics}. Early work \cite{yamada1997} experimentally established force thresholds for pain. A line of investigation \cite{haddadin2009a, haddadin2009, haddadin2007, behrens2014, laffranchi2009} draws collision metrics from automotive testing standards and experimentally establishes the impact of the robot's reflected inertia, velocity, and impact geometry on these metrics.

Analytically relating robot design to these biomechanic safety limits is challenging.  ISO 15066:2016, which sets guidelines for human-robot collision, uses a two mass, single stiffness model under perfect elastic collision to motivate safe robot speed limits \cite{iso2016}. For low-stiffness flexible-joint robots, it is suggested that link-side dynamics dominate collision force and that motor inertia can be neglected \cite{haddadin2009}. While later work \cite{haddadin2010} elaborates, a wide range of joint stiffness are used in practice, from series-elastic and variable stiffness actuators ($20-1000$ Nm/rad \cite{kong2009a, petit2015}), impedance-controlled flexible-joint robots (e.g. Kuka IIWA $\sim 10$kNm/rad \cite{jubien2014a}), to a standard harmonic drive gearbox ($>$15 kNm/rad), suggesting analysis which can accommodate a range of joint stiffnesses is useful.

For contact with a pure-stiffness environment, there is established literature for stability \cite{kazerooni1990, surdilovic2007, vukobratovic2009}, and limiting the peak force of a rigid, perfectly backdriveable robot \cite{heinzmann2003} using inelastic/elastic collision models. Energy-based planning methods have also been introduced, to bound kinetic energy in robot links \cite{laffranchi2009}, or bound maximum power flow between robot and environment \cite{geravand2016}. 

To quantify the effect of compliance and control on collision force, this paper adapts the Sobolev norm as a system norm, allowing a rigorous bound on maximum collision force with an arbitrary environment. The Sobolev norm is shown to be the $\mathcal{H}_2$ norm of a transformed system, allowing computation with existing methods. This norm is validated experimentally and in simulation, for pure stiffness as well as inertial environments.  The impact of joint and interface compliance, motor dynamics, and control on a linearized two-inertia robot model is investigated. Finally, a trade-off between collision performance and environmental estimation uncertainty is shown. 

\section{System Metrics for Interactive Robots}

This section motivates the collision model and approaches for translating collision metrics into system norms. 

\subsection{Impact Metrics and Models}

Some impact metrics used in experimental literature can be seen in Table \ref{collision_metrics}, where the force and velocity at the interaction port are denoted $f(t)$ and $v(t)$, and acceleration of an environmental inertia is $a(t)$ (e.g. head acceleration). 

Contact is here modelled with a rigidly coupled robot/environment - neglecting unilateral contact conditions. The robot model is seen in Figure \ref{physical_sys}, with two inertias (motor and link inertia), and two stiffnesses (joint and end-effector compliance). This model accommodates impedance and admittance-controlled robots, which respectively measure joint torque at $K_{jt}$ or end-effector force/torque near $K_e$ (wrist force/torque sensor). For a serial manipulator, this model can be considered as a linearization about a contact pose. 

To model the relative velocity $v_0$ between robot and environment at the moment contact is made, a state-space model could be used with initial conditions $\dot{\theta}=\dot{q}=v_0$. Here, it is convenient to represent this initial velocity as a Dirac delta input at each robot inertia, scaled by that inertia's momentum.  This impulse, when applied at $t=0$, induces the initial velocity $v_0$. As any output force is linear in this contact velocity, a unit velocity $v_0=1$ is assumed.

\subsection{Metric Bounds by Norms}
Given the model in Figure \ref{physical_sys}, it is desired to guarantee bounds on the collision metrics seen in Table \ref{collision_metrics} as a function of robot properties. Let the force of interest be $f(t)=K_{e}(x-q)$, the contact force at the interface with the environment. Denote the transfer function from $\delta\rightarrow f$ as $G_f(s)$.  Recall that $max |f(t)| = ||f(t)||_\infty$, and when there is no other input to the system ($x=0$, $\tau_m=0$, $\delta$ is the Dirac impulse), $f(t)=\mathscr{L}^{-1}\left\{G_f(s)\right\}$ \cite{zhou1996}. Although $f(t)$ is, in general, an arbitrary output of the system, it will be here regarded as the force response to the unit velocity initial condition with a specified environment and control policy.

If a different linear collision model is written with input $u(t)\in\mathcal{L}_\infty$ (note that an impulse $\delta\notin\mathcal{L}_\infty$), with transfer function $G_{uf}(s)$ for dynamics $u\rightarrow f$, the force can be bounded as $\Vert f(t) \Vert_\infty < \Vert g_{uf}(t) \Vert_1 \Vert u(t) \Vert_\infty$ \cite{doyle2013}, but the $1$ norm of an impulse response $g_{uf}(t)=\mathcal{L}^{-1}\left\{G_{uf}(s)\right\}$ is not easily computed, and the conservatism of this norm in collision analysis is noted \cite{vukobratovic2009}.

\subsection{Sobolev Spaces}
Recall that while norm equivalence is guaranteed over a finite-dimensional vector space  $\mathcal{V}$ (i.e. $\exists c\,\,s.t.\,\,\left\Vert x\right\Vert _{q}<c\left\Vert x\right\Vert _{p}\,\,\forall x\in\mathcal{V}$), this does not hold for uncountably infinite spaces, such as continuous-time signals  $f(t)\in\mathcal{L}_p$. So, bounding $\Vert f(t) \Vert_2$ or $\Vert f(t) \Vert_1$ does not guarantee a bound on $\Vert f(t) \Vert_\infty$.

If a signal is continuous with a continuous derivative, Sobolev spaces can be used to find absolute bounds on $\Vert f(t) \Vert_\infty$. Define the Sobolev norm as
\begin{eqnarray}
\left\Vert f(t)\right\Vert_{W^{1,p}}&=&\Vert f\left(t\right)\Vert^p_{p}+\Vert \dot{f}\left(t\right)\Vert^p _{p},
\end{eqnarray}
 then the following inequality holds from \cite{brezis2010}, Theorem 8.8,
\begin{eqnarray}
\left\Vert f\left(t\right)\right\Vert _{\infty}\leq \gamma\left\Vert f\left(t\right)\right\Vert _{W^{1,p}}\,\,\, \forall f\in W^{1,p}, \label{sob_norm}
\end{eqnarray}
for a fixed constant $\gamma$. The proof is reproduced in the Appendix, with the intuition that while a signal with a finite square integral can have an arbitrarily large peak (e.g. a Gaussian distribution as covariance approaches zero), if the square integral of a signal and its derivative are bounded, a bound on the absolute signal peak exists.

If the $2-$norm is used and $f(t)$ is scalar, $\left\Vert f\left(t\right)\right\Vert _{W^{1,2}}$ can be expressed in the frequency domain with Parseval's Theorem \cite{zhou1996}, yielding: 
\begin{eqnarray}
\Vert f(t) \Vert_{W^{1,2}} &= \Vert F(j\omega) \Vert^2_2 + \Vert j\omega F(j\omega)\Vert^2_2  \label{sob_norm_bound}
\end{eqnarray}
where $F\left(j\omega\right)$ is the Fourier transform of $f\left(t\right)$. For the collision model here, $F(j\omega) = G_f(j\omega)$, the transfer function from $\delta \rightarrow f$.

When the signal $F(j\omega)$ is a rational polynomial with relative degree two, poles only in the open left half plane, and $\lim_{s\rightarrow0}|F(s)|<\infty$, the Sobolev norm exists (i.e. is finite), and can be easily computed from existing signal theory. 

These existence conditions restrict the signals which can be analyzed with the Sobolev norm, but when the force of interest is the force through a stiffness element between two inertias, this condition is satisfied. As force is typically measured by strain gauges on a pure stiffness (in both joint torque sensors and 6-DOF force/torque sensors), this condition is practical for bounding the measured force on real-world robots.

\section{Mechatronic Design for Collision\label{validation}}
This section validates the Sobolev norm for interactive robots in contact with a pure stiffness, and a stiffness/inertia environment. 

\subsection{Pure Stiffness Environment Validation}
\begin{figure}[h]
	\centering
	\includegraphics[width=.7\columnwidth]{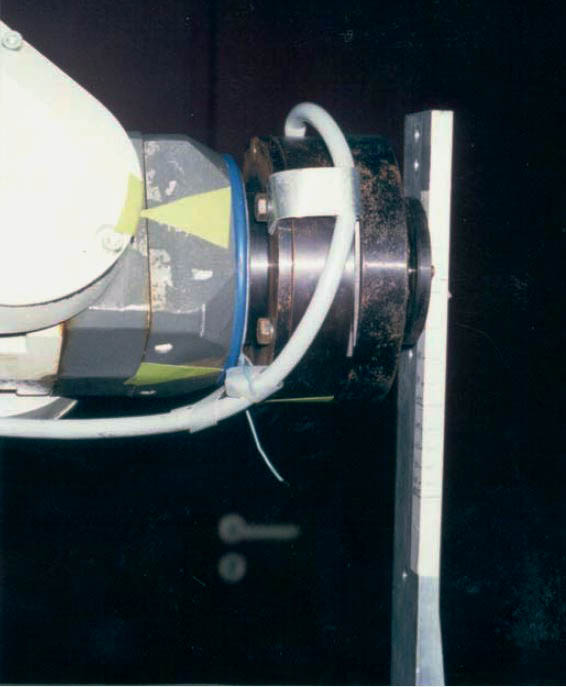}  
	\protect\caption{End-effector of a Manutec r3, in contact with pure-stiffness environment ($\sim 56kN/m$) as used in experimental validation of the proposed collision norm. \label{manut_setup}}
\end{figure}

Contact experiments with a Manutec r3 industrial manipulator under admittance control, as seen in Figure \ref{manut_setup}, with details on controller architecture in \cite{surdilovic2007}. The admittance parameters (stiffness, damping, and inertia) are varied over a set of 64 contact transitions, with ranges shown in Table \ref{sim. params}.
 
The model which the Sobolev norm is applied to is a linearized model for the position-controlled robot, and the outer-loop admittance control parameters (target mass, damping and stiffness). The environment is identified as a $56 kN/m$ stiffness. This gives a total transfer function:
\begin{eqnarray}
    F(s) = \frac{G}{1+CAGK_e+GC+GKe}\left(Mv_0+CX^d\right) \label{adm_tf}
\end{eqnarray}
where $C(s)$ is the position controller, $G(s)$ linearized model from torque to position, $K_e$ environmental stiffness, $A(s)$ the target admittance, $F(s)$ the interaction force and $X^d(s)$ the reference trajectory signal.  Model and controller values used can be seen in Table \ref{sim. params}. The position reference is dropped in this analysis because the steady-state force it induces gives an infinite Sobolev norm, and the transient response (which contains potentially dangerous force peaks) is assumed to be dominated by the momentum of the robot (i.e. the $Mv_0$ term in \eqref{adm_tf}). 

\begin{figure}[h]
	\centering
	\includegraphics[width=\columnwidth]{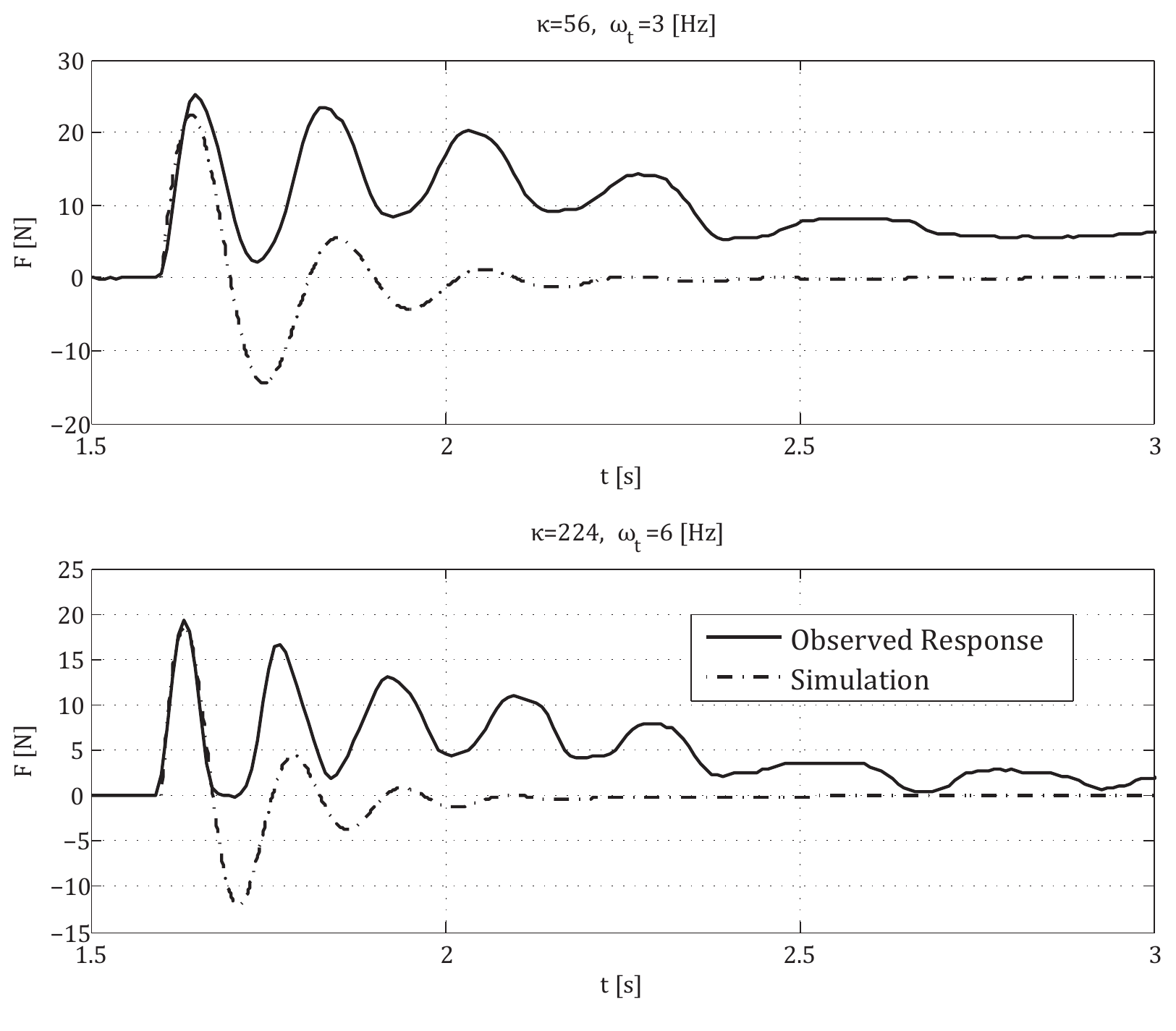}  
	\protect\caption{Force response in two contact transitions compared with simulated response (simulation ignores trajectory reference, leading to steady-state error). \label{time_domain_compare}}
\end{figure}

For each contact experiment, the model was simulated with identical controller parameters, two example experimental and model force trajectories can be seen in Figure \ref{time_domain_compare}. A good correspondence in the initial force peak, resonant frequency, and decay rate are achieved with a model response using the parameters of Table \ref{sim. params}. Steady-state error between model and experiment arises from the contribution of the $X^d$ reference trajectory, which is not included in the simulation.

\begin{figure}[h]
	\centering
	\includegraphics[width=\columnwidth]{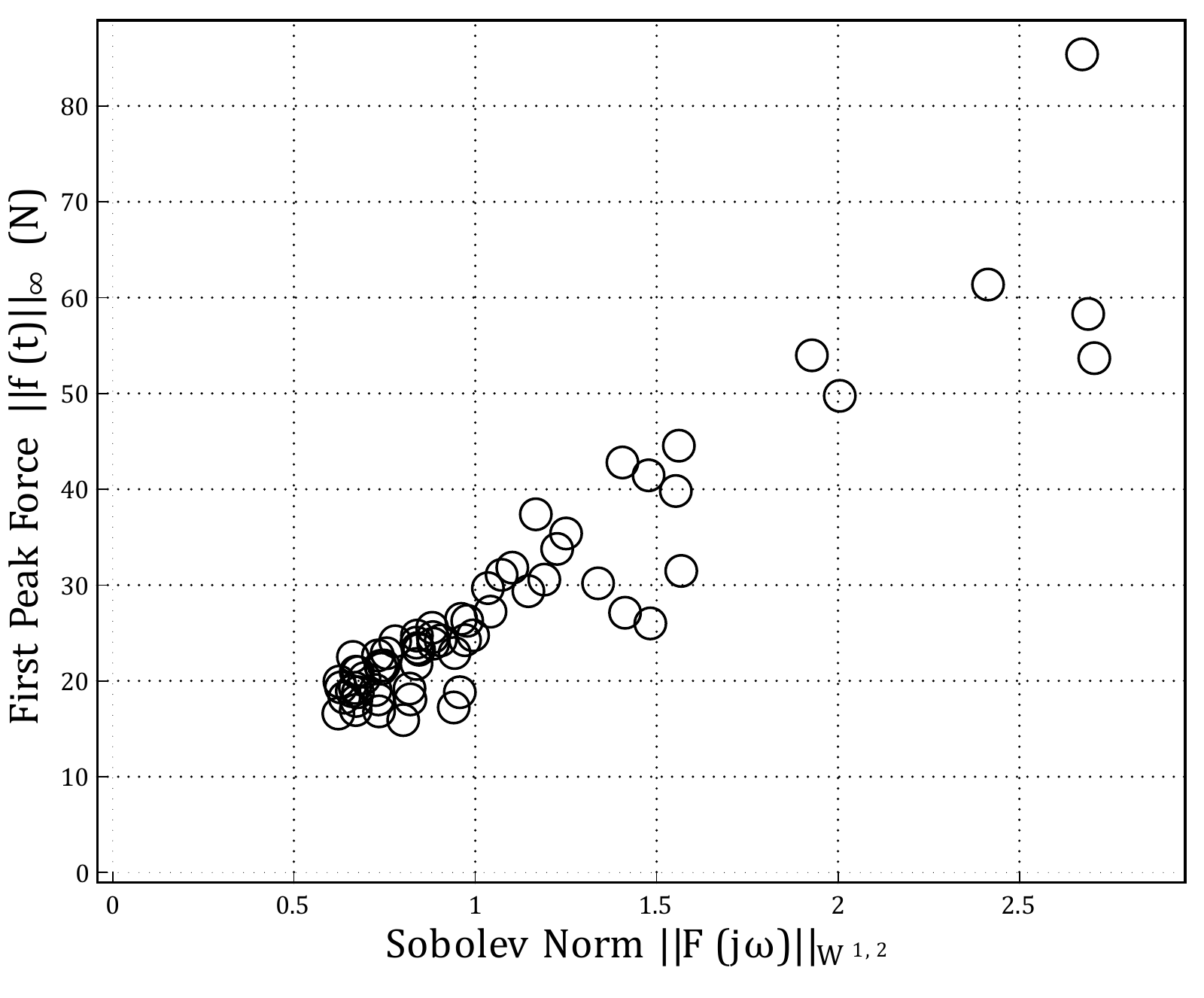}  
	\protect\caption{Peak initial force of the experiment compared with Sobolev Norm of the matching model for various admittance parameters. \label{norm_validation}}
\end{figure}

The model is then used to calculate the $\Vert f(t) \Vert_{W^{1,2}}$ norm for each controller configuration, which is compared with the corresponding observed maximum force from the experiments. Results can be seen in Figure \ref{norm_validation}, achieving good correspondence between the Sobolev Norm and observed peak force across a range of peak force values. 

\begin{table}
	\begin{centering}
			\renewcommand{\arraystretch}{1.3}
		\caption{Model parameters fit to the Manutec r3\label{sim. params}}
		\begin{tabular}{|c|c|l|}
			\hline 
			Param & Value & Units/Notes\tabularnewline
			\hline 
			\hline 
			$G\left(s\right)$ & $\left(308s^{2}+40s\right)^{-1}$ & with $M=308$\tabularnewline
			\hline 
			$C\left(s\right)$ & $85s+370$ & position control bandwidth $\sim4$ Hz\tabularnewline
			\hline 
			$K_{e}$ & $5.6e4$ & $[N/m]$\tabularnewline
			\hline 
			$A\left(s\right)$ & $\frac{1}{M_{t}\left(s^{2}+2\xi\omega_{t}+\omega_{t}^{2}\right)}$ & Ranges used:$\begin{cases}
			\omega_{t} & \left[.5,\,1,\,3,\,6,\,8\right]\\
			M_{t} & \left[25-400\right]
			\end{cases}$ \tabularnewline
			\hline 
		\end{tabular}
		\par\end{centering}
\end{table}

\subsection{Inertial Environment Validation}
To validate the norm in inertial environments, a simulation was used based on Figure \ref{physical_sys}, with the environment being a damped inertia ($f = B_h\dot{x}+I_h\ddot{x}$). The robot parameters are motivated by the first joint of the KUKA LWR as identified in \cite{jubien2014a}, and the human parameters (i.e. $K_e$ and environment inertia $H$) from Table A.3 in  \cite{iso2016}. These human parameters are translated into rotational coordinates under an assumed contact radius of $.2$-$.75$ meters. 
\begin{table}
	\begin{centering}
		\renewcommand{\arraystretch}{1.3}
		\caption{Robot and Human Model Parameters\label{rob_and_hum_param}}
		\begin{tabular}{|c|c|l|}
			\hline 
			Param & Value & Units/Notes\tabularnewline
			\hline 
			\hline 
			$M$ & 3.19 & Kgm\textsuperscript{2}\tabularnewline
			\hline 
			$B$ & 24.3 & Nms/rad\tabularnewline
			\hline 
			$K_{jt}$ & 10 & kNm/rad\tabularnewline
			\hline
			$I$ & 4.5 & Kgm\textsuperscript{2}\tabularnewline
			\hline 
			$V$ & 20.3 & Nms/rad \tabularnewline
			\hline 
			$K_e$ & $\in\{10r^2, 150r^2\}$ & kNm/rad \tabularnewline
			\hline
			$B_h$ & $75r^2$ &  Nms/rad \tabularnewline
			\hline
			$I_h$ & $\in\{.6r^2, 175r^2\}$ & Kgm\textsuperscript{2} \tabularnewline
			\hline
			$r$ & $\in\{.2, .75 \}$ & m \tabularnewline
			\hline
		\end{tabular}
		\par\end{centering}
\end{table}
\begin{figure}[h]
	\centering
	\includegraphics[width=\columnwidth]{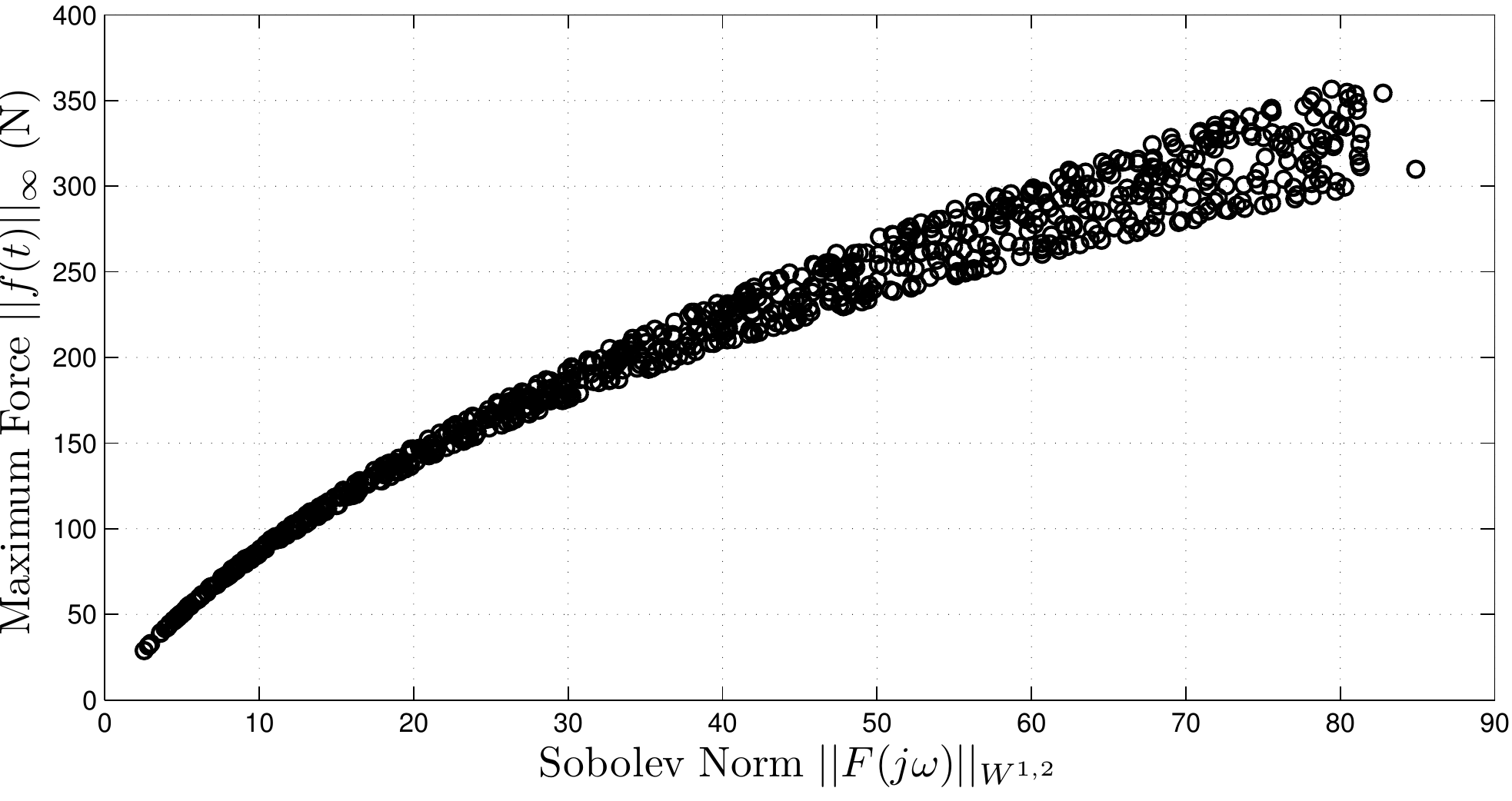}  
	\protect\caption{Norm validation in simulation with inertial environment. \label{inertial_env_sob_norm}}
\end{figure}

The results seen in Figure \ref{inertial_env_sob_norm} compare the maximum force observed in simulated collision with the Sobolev norm of the model. Again a good correspondence is achieved over the range of stiffnesses and inertias expected from humans, with a largely monotonic relationship. Qualitatively, this norm is sensitive to damping (just as the $\mathcal{H}_{\infty}$ norm is), and sufficient damping on the robot model must be included to maintain good predictions.

\subsection{Discussion}
The relationships seen in Figures \ref{norm_validation} and \ref{inertial_env_sob_norm} suggests that although the Sobolev norm is only an upper bound in theory, it can be useful as a design tool: within a family of related systems, decreasing the Sobolev norm reduces peak force.	 However, the experiments here only demonstrate the Sobolev norm on systems where parameters of a fixed architecture are varied. Although this is relevant for mechatronic design (e.g. choosing joint stiffness, interface compliance, controller gains), additional validation is required before using the Sobolev norm to compare impact force between arbitrary systems. 

\section{Design and analysis via Sobolev norm}
With the Sobolev norm providing a useful design approximation for collision performance, its compatibility with established $\mathcal{H}_2$ theory can be used to investigate the impact of joint flexibility, interface compliance, and control on peak collision force.

\subsection{State Space Model}
The system shown in Figure \ref{physical_sys} can be directly written in state-space form, separating the environment by taking it's velocity $\dot{x}$ as input. For a pure-stiffness environment, the stiffness of environment can be incorporated in $K_{e}$ and $\dot{x}=x=0$. The augmented state $\xi=[\theta,\dot{\theta},q,\dot{q},x]$ and the decomposition $B = [B_\delta, B_\tau, B_{\dot{x}}]$ will be used in the following.

\begin{eqnarray}
 \frac{d}{dt}\xi	&=&A\xi+B\left[\begin{array}{c}
 \delta\\
 \tau_{m}\\
 \dot{x}
 \end{array}\right] \\
 w=\left[\begin{array}{c}
 f\label{ss_eqns}\\
 \dot{f}
 \end{array}\right]	&=&C_w\xi+D\left[\begin{array}{c}
 \delta\\
 \tau_{m}\\
 \dot{x}
 \end{array}\right]\nonumber
\end{eqnarray}

\begin{eqnarray*}
\begin{split}
	\small
\left[\begin{array}{c|c}
A & B\\
\hline
C_w & D_w
\end{array}\right]=\\
\setlength\arraycolsep{1pt}\left[\begin{array}{ccccc|ccc}
0 & 1 & 0 & 0 & 0 & 0 & 0 & 0\\
\texttt{-}K_{jt}J^{\texttt{-}1} & \texttt{-}BJ^{\texttt{-}1} & K_{jt}J^{\texttt{-}1} & 0 & 0 & 1 & J^{\texttt{-}1} & 0\\
0 & 0 & 0 & 1 & 0 & 0 & 0 & 0\\
K_{jt}I^{\texttt{-}1} & 0 & \texttt{-}\left(K_{jt}\texttt{+}K_{e}\right)I^{\texttt{-}1} & \texttt{-}VI^{\texttt{-}1} & 0 & 1 & 0 & 0\\
0 & 0 & 0 & 0 & 0 & 0 & 0 & 1\\
\hline
0 & 0 & K_{e} & 0 & \texttt{-}K_{e} & 0 & 0 & 0\\
0 & 0 & 0 & K_{e} & 0 & 0 & 0 & \texttt{-}K_{e}
\end{array}\right]
\end{split} \label{ss_physical_sys}
\end{eqnarray*}

\subsection{Sobolev as an $\mathcal{H}_2$ norm \label{sob_norm_calc}}
The Sobolev norm of the force response of the impulse collision model can be written as
\begin{eqnarray}
 \Vert f \Vert_{W^{1,2}} &= &\Vert f(t) \Vert_2^2 + \Vert\dot{f}(t) \Vert_2^2 \\
 & = &\int_0^\infty w^T(t)w(t) dt  \\
 & =&\Vert G_{\delta f}(s) \Vert_2^2, 
\end{eqnarray}
where $G_{\delta f}$ is the transfer function from $\delta$ to $f$. The $\mathcal{H}_2$ norm can be directly calculated through the controllability or observability Gramian \cite{zhou1996}, for example
\begin{eqnarray}
\Vert G_{\delta w} \Vert_2^2 &=& \mathrm{trace}\left(B_\delta^TL_oB_\delta\right)\\
A^TL_o + L_oA &=& -C_w^TC_w.
\end{eqnarray}

The state feedback controller $\tau_m=K\xi$ which minimizes the sum of the square of the Sobolev norm and a quadratic control penalty can be found through an LQR problem \cite{zhou1996}
\begin{eqnarray}
K_{lqr} &= \arg\min_{K} \int_0^\infty w^T(t)w(t)+\tau_m(t)^TR\tau_m(t)dt \label{lqr_gains}\\ 
& = \arg\min_K \Vert f(t) \Vert^2_{W^{1,2}} + \int_0^\infty\tau_m(t)^TR\tau_m(t)dt,
\end{eqnarray}
with minimum cost of $\mathrm{tr}(B_\delta^TSB_{\delta})$, where $S$ is the unique solution to the LQR Ricatti Equation
\begin{eqnarray}
A^TS+SA-SB_\tau R^{-1}B_\tau^TS+C_w^TC_w=0.
\end{eqnarray}

Although the LQR controller is not practical (complete state-space information is rarely available, high-DOF robots are not approximately linear, robot control must also achieve other objectives), this gives a baseline to compare the possible impact of control on collision.

\subsection{Joint Stiffness Analysis}
\begin{figure}[h]
	\centering
	\includegraphics[width=\columnwidth]{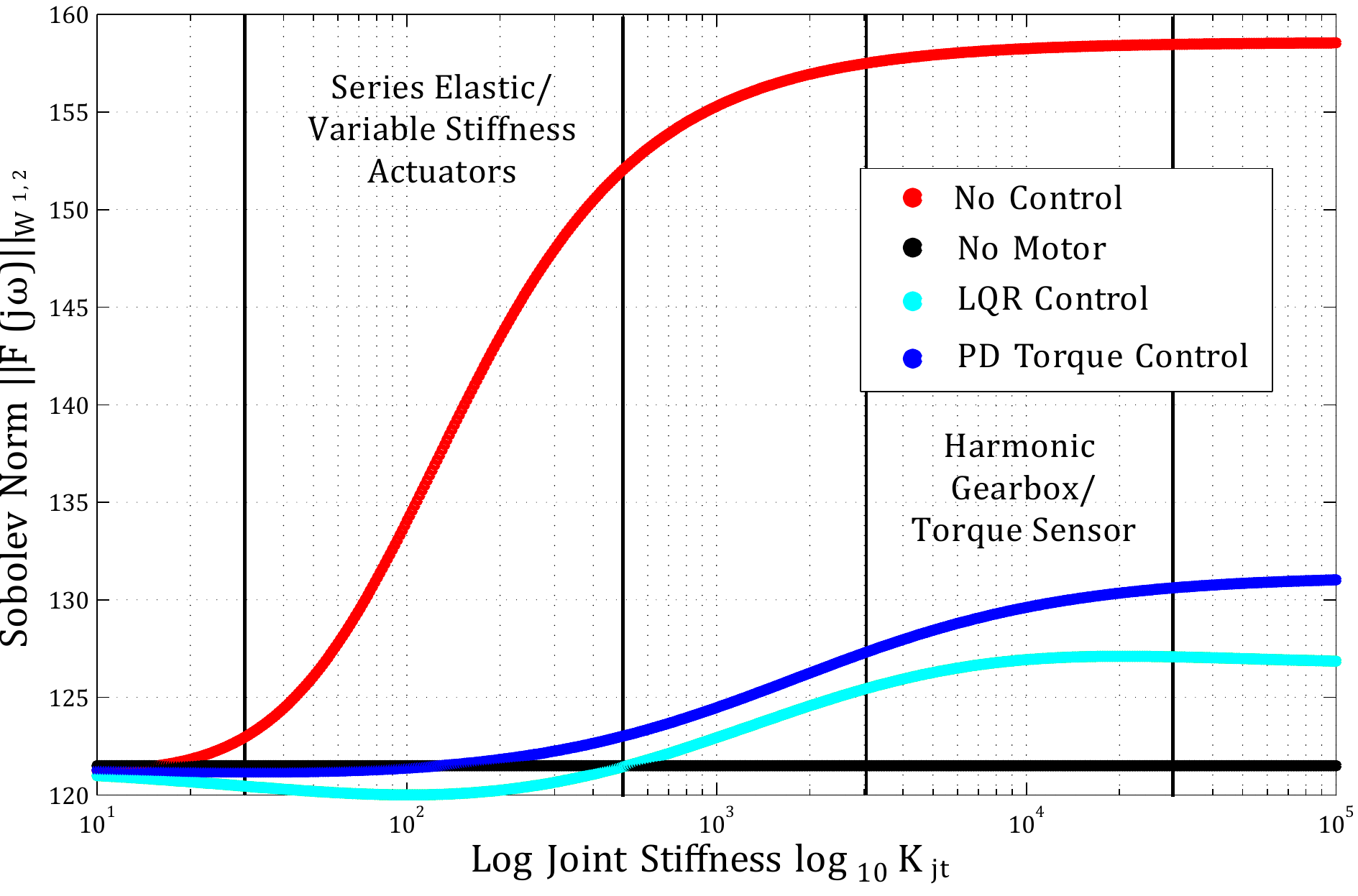}  
	\protect\caption{Collision performance of robot with no control ($\tau_m=0$), no motor dynamics ($J=B=0$), torque control and LQR control, with parameters from Table \ref{rob_and_hum_param} at $K_e=30$ kNm/rad. Note $y$ axis range. \label{collision_perf_comparison}}
\end{figure}

The impact of control and joint flexibility on collision performance can now be investigated. The complete model is the system in Figure \ref{physical_sys}, under parameters in Table \ref{rob_and_hum_param}, in contact with a $K_e=30$ kNm/rad environment. Four different cases can be compared: the uncontrolled system, the system without motor dynamics omitted (i.e. $J=0, B=0$), the system with torque control $\tau_m = (K_p+K_ds)K_{jt}(\theta-q)$ ($K_p = 1.5$, $K_d = .2$), and the system under state feedback with LQR gains from \eqref{lqr_gains} with $R=5$. These four cases are compared as the joint stiffness changes in Figure \ref{collision_perf_comparison}.

At low joint stiffnesses, control and motor dynamics make no difference, and all systems have the same collision performance which is dominated by link dynamics. As joint stiffness increases, the no control system collision performance increases to $150\%$ of the no motor case, showing how at higher joint stiffnesses, the motor dynamics affect the collision performance. However, the effect of motor dynamics is substantially reduced by the inner-loop torque control (for both PD or LQR). Although physical compliance reduces the impact of motor dynamics, it does not need to be very low to achieve good collision performance if control is used.

LQR and PD torque control present similar behaviors, which can be seen from the similarity of their expressions. Typical LQR results with $R=5 \rightarrow K_{lqr}=[1.28,\, .03,\, -1.30,\, -.05,\, 0]1e4$ and a PD torque controller written as a similar state feedback gain $K_{pd} = [Kp,\, Kd,\, -Kp,\, -Kd,\,0]K_{jt}$. The matching of signs (as well as approximate magnitude commonly used on PD control of joint torque) shows that inner-loop PD torque control is well-aligned with reducing peak impact force.

\subsection{Interface Stiffness Analysis}
While decreasing the stiffness of the interface $K_e$ provides improvements in collision performance, there is a trade-off with performance. Classical interpretations of performance in control theory is the ability to reshape the apparent dynamics of the system (i.e. reshaping the transfer functions of disturbance rejection, reference tracking, noise sensitivity). For interactive robots, the apparent dynamics are a direct performance goal as well, although the range of dynamics required is not as easily formalized. 

There is a secondary sense of performance for interactive robots; their ability to discover actionable information from the environment. Just as actuator dynamics and non-collocation limit the dynamics which can be rendered, sensor dynamics and non-collocation limit the information flow from the environment. As a concrete example, take a robot coming into contact with a kinematic plane at an uncertain position. While knowledge of this plane's location may be important for subsequent behavior, the compliance at the end-effector can limit the ability to estimate the hidden information of the environment. 

Working this example within the model in Figure \ref{physical_sys} as written in \eqref{ss_physical_sys}, this information transfer can be stated as the covariance of the posterior estimate $p(\hat{x}|y(0:T))$, where $y(0:T)$ is the history of measurements available. Impedance- and admittance-controlled  robots use motor position measurements, along with measurements of the force along $K_{jt}$ and $K_{int}$, respectively which can be written as emission equations on the states seen in \eqref{ss_physical_sys}.
\begin{eqnarray}
y_{imp}&=&\left[\begin{array}{ccccc}
1 & 0 & 0 & 0 & 0\\
K_{jt} & 0 & -K_{jt} & 0 & 0
\end{array}\right]\xi\\
y_{adm}&=&\left[\begin{array}{ccccc}
1 & 0 & 0 & 0 & 0\\
0 & 0 & K_{int} & 0 & -K_{int}
\end{array}\right]\xi
\end{eqnarray}

\begin{figure}[h]
	\centering
	\includegraphics[width=\columnwidth]{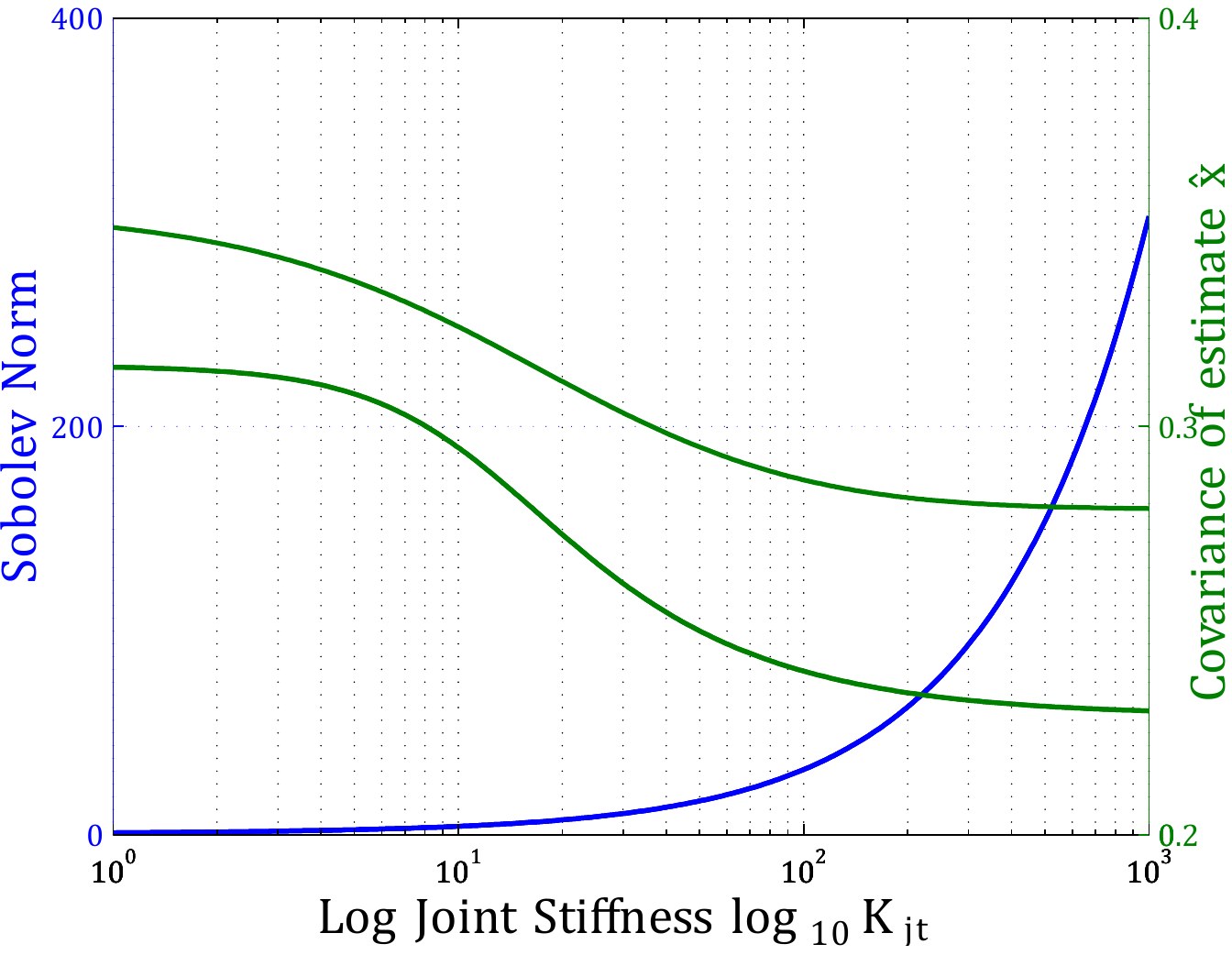}  
	\protect\caption{Effect of interface stiffness $K_{e}$ on collision performance (blue) and covariance of estimate of environment location under impedance (top, green) and admittance (bottom, green), parameters from Table \ref{rob_and_hum_param} at $K_{jt}=10$ kNm/rad. \label{info_transfer}}
\end{figure}

Under a unified noise model, with process noise covariance $\Sigma_w = \mathtt{diag}([.1, .5, .1, 10, 15])$, sensor noise $\Sigma_v = \mathtt{diag}([.1 10])$, the steady-state covariance of the estimate of environment location $\hat{x}$ can be found in closed form from standard solutions to the Kalman filter. The impact of $K_{int}$ on this covariance can be seen in Figure \ref{info_transfer}, where it is compared with collision performance. Collision performance monotonically improves as stiffness decreases, but the quality of the estimation of environment decreases at lower stiffnesses as well.   Low stiffnesses allow torque-sensor noise to dominate the force arising from the environment.

\section{Conclusion}
This paper introduced a system norm for collision analysis based on the Sobolev norm. By bounding the maximum force on a pure stiffness, it provides useful predictions for human-robot and robot-environment contact transitions. The norm was validated with experiments of a pure stiffness environment, and simulation of an inertial environment, and the maximum impact force and Sobolev norm are shown to correspond well over parametric variation of the systems. The Sobolev norm of a system is then stated as the $\mathcal{H}_2$ norm of a related system, and existing control theory used to establish the impact of joint stiffness and interface stiffness on impact performance, towards validating design approximations which are used in practice.

This norm provides the advantage over existing methodologies (perfect elastic collision of two inertias with single stiffness) by allowing the accounting for joint stiffness, motor dynamics, and control. Furthermore, collision can be considered over a wider continuum of environments, also supporting design for contact transitions with industrial, high-stiffness environments. 

\section*{Appendix}
The proof of \cite{brezis2010}, Theorem 8.8 is recreated with additional steps for clarity. Let $f\left(t\right)\in C^{1}$, a continuous function with a continuous derivative, and for $1\leq p<\infty$ define the function $G\left(s\right)=\left|s\right|^{p-1}s$. The function $v=G\left(f\right)$ is in $C^{1}$ and
\begin{align*}
v' & =G'\left(f\right)f'\\
& =p\left|f\right|^{p-1}f'\\
&\Rightarrow v=\int_{-\infty}^{t}p\left|f\left(\tau\right)\right|^{p-1}f'\left(\tau\right)d\tau.
\end{align*}
Note that $\left|v\left(t\right)\right|=\left|f\left(t\right)\right|^{p}$, and can be bounded as

\begin{align*}
\left|v\left(t\right)\right| & \leq\int_{-\infty}^{t}p\left|f\left(\tau\right)\right|^{p-1}\left|f'\left(\tau\right)\right|d\tau,\\
&\Rightarrow  \left|f\left(t\right)\right|^{p}\leq p\left\Vert f\right\Vert _{p}^{p-1}\left\Vert f'\right\Vert _{p}.
\end{align*}
Manipulation of this term with a Sobolev norm $\left\Vert f\right\Vert _{W^{1,2}}=\left\Vert f\right\Vert _{p}^{p}+\left\Vert f'\right\Vert _{p}^{p}=\alpha^{p}$, and noting $p^{1/p}\leq e^{1/e}\,\,\forall p>1$ gives
\[
\left|f\left(t\right)\right|\leq \gamma\alpha
\]
for a positive constant $\gamma$ which does not depend on $f(t)$.
\bibliographystyle{IEEEtran}
\bibliography{lib}

\end{document}